\documentclass[a4paper,10pt,twocolumn,english]{article}

\usepackage{babel}
\usepackage[utf8]{inputenc}
\usepackage[T1]{fontenc}

\usepackage{amsfonts,amssymb,amsmath,amsthm}
\usepackage{mathtools} 
\usepackage{subfigure}
\usepackage{graphicx}
\usepackage[footnotesize]{caption}
\usepackage{bm}
\usepackage[table,xcdraw]{xcolor}
\usepackage{multirow}
\usepackage{url}

\newcommand{\bs}{\boldsymbol}
\newcommand{\cl}{\mathcal}
\newcommand{\ie}{\emph{i.e.}, }
\newcommand{\eg}{\emph{e.g.}, }
\newcommand{\bb}{\mathbb}
\DeclareMathOperator{\iid}{iid}
\newcommand{\distiid}{\sim_{\iid}} 
\renewcommand{\Vec}[1]{\bm{#1}} 
\newcommand{\expec}[1]{\mathop{{}\mathbb{E}}_{#1}} 
\DeclarePairedDelimiterX{\norm}[1]{\lVert}{\rVert}{#1} 

\newcommand{\ts}{\textstyle}
\newcommand{\ssum}[1]{\underset{#1}{\sum}}

\usepackage[top=1.5cm, left=1.5cm, right=1.5cm, bottom=1.5cm]{geometry}

\title{Compressive Classification (Machine Learning without learning)}

\author{Vincent Schellekens$^*$ and Laurent Jacques\thanks{E-mail: {\em \{vincent.schellekens,~laurent.jacques\}@uclouvain.be}. ISPGroup, ELEN/ICTEAM, UCLouvain (UCL), B1348 Louvain-la-Neuve, Belgium. VS and LJ are funded by Belgian National Science Foundation
(F.R.S.-FNRS).}}

\renewenvironment{abstract}{\bf\small {\em\ Abstract---}}{}

\date{}

\begin{document}

\maketitle

\begin{abstract}Compressive learning is a framework where (so far unsupervised) learning tasks use not the entire dataset but a compressed summary (sketch) of it. We propose a compressive learning classification method, and a novel sketch function for images.
\end{abstract}

\vspace{-1mm}

\section{Introduction and background}
\label{sec:introduction}
Machine Learning (ML)---inferring models from datasets of numerous learning examples---recently showed unparalleled success on a wide variety of problems. However, modern massive datasets necessitate a long training time and large memory storage. The recent \textit{Compressive Learning} (CL) framework alleviates those drawbacks by computing a compressed summary of the dataset---its \emph{sketch}---prior to any learning~\cite{gribonval2017compressiveStatisticalLearning}.
The sketch is easily computed in a single parallelizable pass, and its required size (to capture enough information for successful learning) does not grow with the number of examples: CLs time and memory requirements are thus unaffected by the dataset size.

\begin{figure}[]
	\centering
	\includegraphics[width=0.99\linewidth]{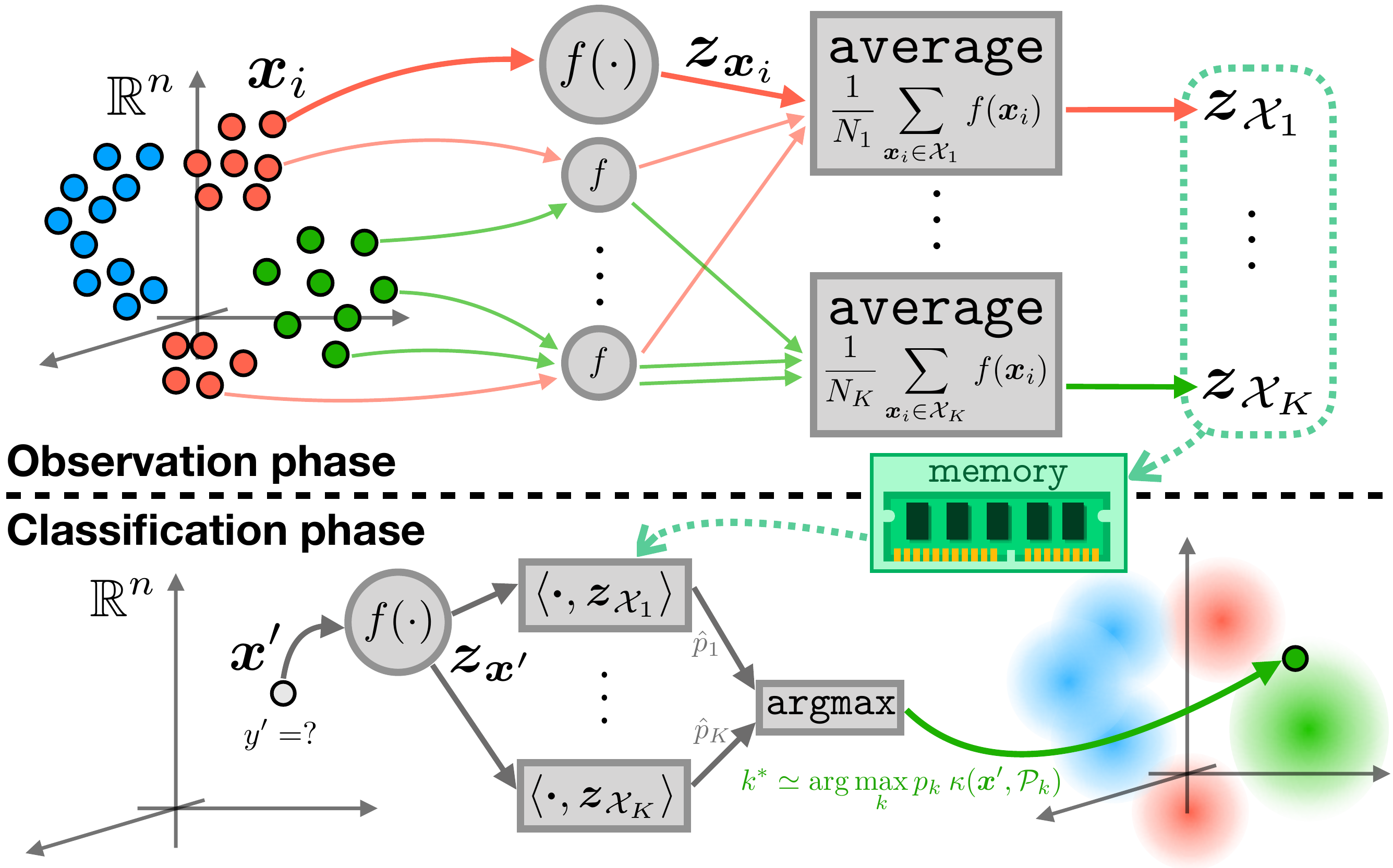} 
	\caption{\textbf{Observation phase:} we record only a summary of the dataset $\cl X$ as the $K$ class sketches $\Vec{z}_{\cl X_k}$: the class average of non-linear maps $\Vec{z}_{\Vec{x}_i} = f(\Vec{x}_i)$ of the examples $\Vec{x}_i$. \textbf{Classification phase:} a new sample $\Vec{x}'$ gets the class label $k^*$ that maximizes the correlation between its sketch $\Vec{z}_{\Vec{x}'}$ and the stored class sketches; this can be interpreted as a MAP classifier in a RKHS $\cl H_{\kappa}$.}
	\label{fig:intro}
	\vspace{-2.0mm}
\end{figure}

So far, CL focused on \textit{unsupervised} ML tasks, where learning examples don't belong to a (known) class~\cite{gribonval2017compressiveStatisticalLearning,keriven2016compressive,keriven2016GMMestimation}. We show that CL easily extends to \textit{supervised} ML tasks by proposing (Sec.~\ref{sec:proposedmethod}) and experimentally validating (Sec.~\ref{sec:experiments}) a first simple \textit{compressive classification method} using only a sketch of the labeled dataset (Fig.~\ref{fig:intro}).
We also introduce a sketch feature function leveraging a random convolutional neural network to better capture information in images. While not as accurate as ML methods learning from the full dataset, this compressive classification scheme still attains remarkable accuracy considering its \textit{unlearned nature}. Our method also enjoys from a nice geometric interpretation, \ie Maximum A Posteriori classification performed in the Reproducible Kernel Hilbert Space associated with the sketch. 

\noindent \textbf{(Unsupervised) Compressive Learning:} Unsupervised ML usually amount to estimate parameters of a distribution $\cl P$, from a dataset ${\cl X := \{\Vec{x}_i \distiid \cl P\}_{i = 1}^{N}} \subset \mathbb R^n$ of examples---associated to an \textit{empirical distribution} $\hat{\cl{P}}_{\cl X} := \frac{1}{N} \sum_{\Vec{x}_i \in {\cl X}} \delta_{\Vec{x}_i}$, with $\delta_{\Vec{u}}$ the Dirac measure at $\Vec{u}$. While most unsupervised ML algorithms require (often multiple times) access to the entire dataset $\cl X$, CL algorithms require only access to the sketch: a single vector $\Vec{z}_{\cl X} \in \bb C^m$ summarizing $\cl X$. This \textit{dataset sketch} $\Vec{z}_{\cl X}$ actually serves as a proxy for the true \textit{distribution sketch} $\cl A(\cl P)$, \ie a linear embedding of the ``infinite-dimensional'' probability distribution $\cl P$ into $\bb C^m$, a space of lower dimension: 
\begin{equation}
\label{eq:sketchDef}
\cl A(\cl P) :=\!\! \expec{\Vec{x} \sim \cl P} f(\Vec{x}) \simeq \Vec{z}_{\cl X} := \cl A(\hat{\cl P}_{\cl X}) = \ts \tfrac{1}{N} \ssum{\Vec{x}_i \in {\cl X}} f(\Vec{x}_i), 
\end{equation}
\noindent where $f$ is a random nonlinear feature map to $\bb C^m$. This map defines a positive definite kernel $\kappa(\Vec{u},\Vec{v}) \! := \! \expec{}\langle f(\Vec{u}),f(\Vec{v})\rangle$, and $\kappa$ in turn provides a Reproducible Kernel Hilbert Space (RKHS) $\cl H_{\kappa}$ to embed distributions; $\cl A$ indirectly maps $\cl P$ to its \textit{Mean Map} $\kappa(\cdot,\cl P) := \expec{\Vec{x} \sim \cl P} \kappa(\cdot, \Vec{x}) \in \cl H_{\kappa}$~\cite{aronszajn1950rkhsThm,smola2007hilbert,sriperumbudur2010hilbertEmbedding}. Existing methods~\cite{keriven2016compressive,keriven2016GMMestimation} use Random Fourier Features~\cite{rahimi2008RFF} as map $f$:
\begin{equation}
\label{eq:f-RFF}
\ts f_{\texttt{RFF}}(\Vec{x}) = \left[ \exp(\mathrm{i}\,\Vec{\omega}_j^T\Vec{x}) \right]_{j=1}^m \quad \text{with} \quad \Vec{\omega}_j \distiid \Lambda,
\end{equation}
and $\kappa$ is then shift-invariant and the Fourier transform of the distribution $\Lambda$: $\kappa(\Vec{x},\Vec{x}') = \varpi(\Vec{x}-\Vec{x}') := (\cl F \Lambda)(\Vec{x}-\Vec{x}')$~\cite{rudin1962bochnerBook}. CL is promising because the sketch $\Vec{z}_{\cl X}$ retains sufficient information (to compete with traditional ML) whenever its size $m$ exceeds some value independent on the number of examples $N$, yielding algorithms that scale well when $N$ increases.

\noindent \textbf{Random Convolutional Neural Networks (CNN):}
Shift-invariant kernels are not that relevant when dealing with images (they are sensitive to image translations for example). Recent studies have shown that the last layer of a \textit{randomly weighted} (convolutional) neural network \texttt{CNN} (combining convolutions with random weights, nonlinear activations, and pooling operations) captures surprisingly meaningful image features~\cite{ulyanov2017deepimageprior,giryes2016deeprandom,rosenfeld2018randomnet,cho2009kernel}. We thus propose the feature map $f_{\texttt{CNN}}(\Vec{x}) = \texttt{CNN}(\Vec{x}) \in \bb R^m$ as sketch map $f$ for images: the associated kernel $\kappa$ is (for a fully connected network) an \textit{arc-cosine kernel}, that surpasses shift-invariant kernels for solving image classification tasks with kernel methods~\cite{cho2009kernel}.

\section{Compressive learning classification}
\label{sec:proposedmethod}

\textbf{Observation phase:} \textit{Supervised} ML infers a mathematical model from a labeled dataset $\cl X := \{(\Vec{x}_i,y_i)\}_{i = 1}^N$ where each signal $\Vec{x}_i \in \bb R^n$ belongs to a class $\cl C_k$ as designated by its \textit{class label} $y_i \in [K]$. 
Denoting $p_k := \bb P(\bs x \in \cl C_k) = \bb P(y = k)$, the signals are assumed drawn from an unknown density $\cl P$:\vspace{-1.5mm}
\begin{equation}
\label{eq:modelDensity}
\Vec{x_i} \distiid \ts \cl P = \sum_{k=1}^{K}\ p_k\,p(\Vec{x}|\, \bs x \in \cl C_k) =: \sum_{k} p_k \cl P_k(\Vec{x}).\vspace{-1.5mm}
\end{equation}
As illustrated in Fig.~\ref{fig:intro}(top), our \textit{supervised compressive learning} framework considers that $\cl X$ is not explicitly available but compressed as a collection of $K$ \textit{class sketches} $\Vec{z}_{\cl X_k}$ defined as:\vspace{-1.5mm}
\begin{equation}
\label{eq:classSketches}
\ts \Vec{z}_{\cl X_k} = \mathcal{A}(\hat{\mathcal{P}}_{\cl X_k}) \quad \text{where} \quad \cl X_k := \{ \Vec{x}_i \in \cl C_k \}.\vspace{-1.5mm}
\end{equation}
We can also require approximated \textit{a priori} class probabilities $\hat{p}_k$, \eg $\hat{p}_k = \frac{N_k}{N}$ if we count the class occurrences $N_k = |\cl X_k|$, or setting an uniform prior $\hat{p}_k = \frac{1}{K}$ otherwise.

\noindent \textbf{Classification phase:} Under (\ref{eq:modelDensity}), the optimal classifier (minimal error probability) for a test example $\Vec{x}'$ is the Maximum A Posteriori (MAP) estimator $k^{\texttt{MAP}} := \arg \max_k p_k \cl P_k( \Vec{x}' )$, where $\cl P_k$ is generally hard to estimate. In our CL framework, we classify $\Vec{x}'$ from $\Vec{z}_{\cl X_k}$ and $\hat p_k$ only (Fig.~\ref{fig:intro}, bottom): we acquire its sketch $\Vec{z}_{\Vec{x}'} = f(\Vec{x}')$ and maximize the correlation with the class sketch weighted by $\hat{p}_k$, \ie we assign to $\Vec{x}'$ the label\vspace{-1.5mm}
\begin{equation}
\ts k^* := \arg \max_k \: \hat{p}_k \langle \Vec{z}_{\Vec{x}'},  \Vec{z}_{\cl X_k}\rangle \tag*{(CC)} \vspace{-1.5mm}
\label{eq:classifier}
\end{equation}
Note that this Compressive Classifier (CC) does not require parameter tuning. Interestingly, under a few approximations, this procedure can be seen as a MAP estimator in the RKHS $\cl H_{\kappa}$.  Indeed, we first note that if $m$ is large, the law of large numbers (LLN) provides the \textit{kernel approximation} (KA)  \vspace{-1.2mm}
\begin{equation}
  \label{eq:kern-approx}
\ts \langle f(\bs u),f(\bs v) \rangle \simeq \kappa(\bs u, \bs v),\quad \forall \bs u, \bs v \in \bb R^n.  
\tag*{(KA)}\vspace{-1.2mm}
\end{equation}
Assuming $N_k$ is also large, another use of the LLN gives the \textit{mean map approximation} (MMA): we have both $\hat{p}_k \simeq p_k$ and\vspace{-1.1mm}
\begin{align}
&\ts \ts \langle \Vec{z}_{\bs u},  \Vec{z}_{\cl X_k} \rangle = \frac{1}{N_k} \ssum{\Vec{x}_i \in \cl X_k} \langle f(\bs u),f(\bs x_i)\rangle \underset{\rm \ref{eq:kern-approx}}{\simeq} \frac{1}{N_k} \ssum{\Vec{x}_i \in \cl X_k} \kappa(\bs u,\Vec{x}_i) \nonumber\\
&\label{eq:mean-map-approximation}
\ts \simeq\ \expec{\Vec{x} \sim \cl P_k } \kappa(\bs u,\Vec{x})\ =:\ \kappa(\bs u, \cl P_k) \quad \forall  \bs u \in \bb R^n.
\tag*{(MMA)}\vspace{-1.1mm}
\end{align}

\noindent Consequently, under the KA and MMA approximations,\vspace{-1.5mm}
\begin{equation}
\label{eq:apprMAP}
\ts k^*\ \simeq\ \arg \max_k\ p_k\,\kappa(\bs x', \cl P_k),\vspace{-1.5mm}
\end{equation}
or in other words, we replace $\cl P_k$ in the MAP estimator by its Mean Map $\kappa(\cdot, \cl P_k)$---its embedding in $\cl H_{\kappa}$---such that CC computes a \textit{MAP estimation inside the RKHS} $\cl H_{\kappa}$. In all generality $\kappa(\cdot, \cl P_k)$ is not a probability density function, but can be interpreted as a smoothing of $\cl P_k$ by convolution with $\varpi(\Vec{u}) := \kappa(\Vec{u},0) $ if $\kappa$ is a properly scaled shift-invariant kernel. 
Alternatively, (\ref{eq:apprMAP}) can be seen as a Parzen-windows classifier---a nonparametric Support Vector Machine (without weights learning)---evaluated compressively thanks to the sketch~\cite{duda1973pattern,scholkopf2011learning}.

\section{Experimental proof of concept}
\label{sec:experiments}

\begin{figure}[]
	\centering
	\includegraphics[width=1.0\linewidth]{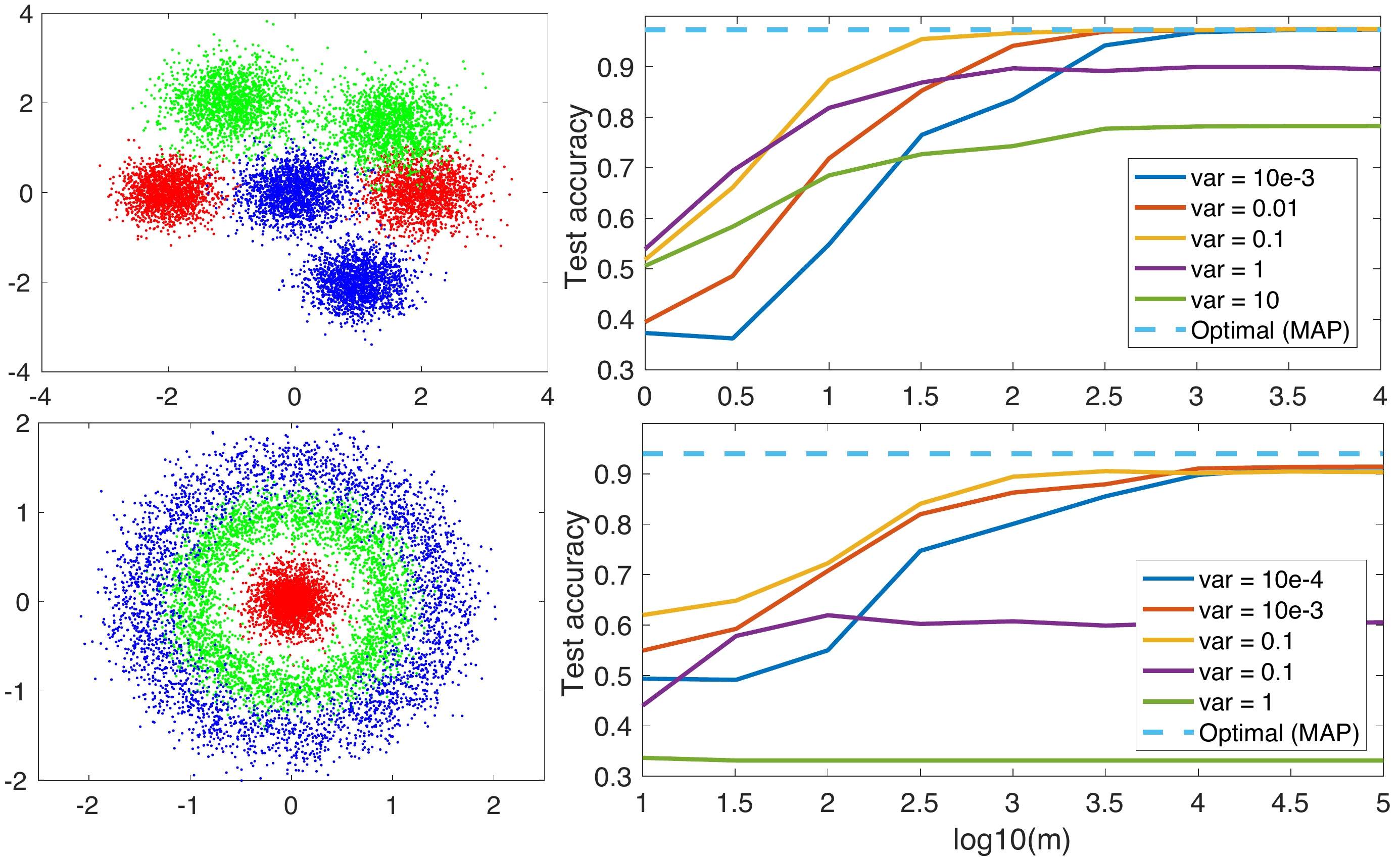} 
	\caption{\textbf{Left}: synthetic $2$-d datasets of $N = 10^4$ examples from $K = 3$ equiprobable classes, separated into $2/3$ for ``training'' (observation phase) and $1/3$ for testing (classification phase). \textbf{Right}: testing accuracy (average over 10 trials) of our compressive classification method for different values of $\sigma$ (noted \texttt{var}) and increasing $m$ (solid), compared to MAP classification (dashed).}
	\label{fig:exp-synth}
	\vspace{-2.0mm}
\end{figure}

\noindent \textbf{Synthetic datasets}: 
We build two datasets that are not linearly separable (Fig.~\ref{fig:exp-synth} left), and sketch them using $f = f_{\texttt{RFF}}$ with $\Lambda \sim \cl N(\Vec{0}, \frac{I_n}{\sigma^{-2}})$: therefore $\kappa(\Vec{u},\Vec{v}) \propto \exp(-\frac{\|\Vec{u}-\Vec{v}\|^2}{2\sigma^2})$. As shown Fig.~\ref{fig:exp-synth}(right), the test accuracy of CC improves with $m$ until reaching---when the KA is good enough---a constant floor depending on the compatibility between $\kappa$ and $\cl P$. Accuracy is almost optimal when $\kappa$ is close to the constituents of $\cl P$ (\eg $1^{\mathrm{st}}$ dataset, $\sigma = 0.1$), but degrades when the kernel scale and/or shape mismatches the data clusters (\eg $1^{\mathrm{st}}$ dataset, $\sigma = 10$; or $2^{\mathrm{nd}}$ dataset). CC thus reaches good accuracy provided $m$ is large enough and $\kappa$ is well adapted to the task.

\noindent \textbf{Standard datasets}: 
We also test CC on some well-known ``real-life'' datasets from the UCI ML Repository~\cite{UCIrepository}. Table~\ref{tab:exp-dataset} compares the error rates of CC and SVM, a fully learned approach. Although worse than SVM, CC is surprisingly accurate considering its compressive nature, low computational cost (especially when $m=50$), and that $\kappa$ is a basic, non-tuned kernel.

\begin{table}[h]
\footnotesize
	\centering
        \scalebox{0.8}{
	\begin{tabular}{l|l|l|l|l|l|l|}
		\cline{2-7}
		& N                       & n                    & K                   & SVM                      & $m = 50$                   & $m = 1000$                 \\ \hline
		\multicolumn{1}{|l|}{}                                &                         &                      &                     &    $2.00 $    &    $6.51 \pm 1.81$       &  $5.51 \pm 1.23$     \\
		\multicolumn{1}{|l|}{\multirow{-2}{*}{Iris}}          & \multirow{-2}{*}{150} & \multirow{-2}{*}{4}   & \multirow{-2}{*}{3} & \cellcolor[HTML]{C0C0C0}$4.00$ & \cellcolor[HTML]{C0C0C0}$8.22 \pm 3.25$ & \cellcolor[HTML]{C0C0C0}$6.18 \pm 2.40$ \\ \hline
		\multicolumn{1}{|l|}{}                                &                         &                      &  &  $0.84$  & $4.56 \pm 2.34$ & $2.43 \pm 0.72$ \\
		\multicolumn{1}{|l|}{\multirow{-2}{*}{Wine}}          & \multirow{-2}{*}{178} & \multirow{-2}{*}{13}   & \multirow{-2}{*}{3}  & \cellcolor[HTML]{C0C0C0}$1.69$ & \cellcolor[HTML]{C0C0C0}$13.75 \pm 4.09$ & \cellcolor[HTML]{C0C0C0}$8.19 \pm 1.29$ \\ \hline
		\multicolumn{1}{|l|}{}                                &                         &                      &                     & $3.67$ & $7.00 \pm 1.40$ & $3.93 \pm 0.39$ \\
		\multicolumn{1}{|l|}{\multirow{-2}{*}{Breast cancer}} & \multirow{-2}{*}{569}   & \multirow{-2}{*}{30} & \multirow{-2}{*}{2} & \cellcolor[HTML]{C0C0C0}$2.13$ & \cellcolor[HTML]{C0C0C0}$9.22 \pm 2.33$ & \cellcolor[HTML]{C0C0C0}$6.23 \pm 0.69$ \\ \hline
		\multicolumn{1}{|l|}{}                                &                         &                      &                     &   $21.03$ & $23.88 \pm 4.37$ & $23.11 \pm 1.05$ \\
		\multicolumn{1}{|l|}{\multirow{-2}{*}{Adult (3 attr.)}}         & \multirow{-2}{*}{30718}      & \multirow{-2}{*}{3}   & \multirow{-2}{*}{2} & \cellcolor[HTML]{C0C0C0}$21.06$ & \cellcolor[HTML]{C0C0C0}$ 36.09 \pm 6.67$ & \cellcolor[HTML]{C0C0C0}$ 35.04 \pm 1.63 $ \\ \cline{1-1} \cline{2-7} 
	\end{tabular}}
	\caption{Standard datasets: train set (white, 2/3 of data) and test set (gray) average error rates $\pm$ standard deviation (in $\%$, 100 repetitions), for SVM and CC with $m \in \{50,1000\}$, and with $\sigma = 2$ (data re-scaled inside $[-1,+1]^n$). \label{tab:exp-dataset}}
	\vspace{-8mm}
\end{table}

\noindent \textbf{Image classification}: 
More challenging are image classification datasets: handwritten digit recognition (MNIST~\cite{MNISTdataset}) and vehicle/animal recognition (CIFAR-10~\cite{CIFARdataset}).
We use $f = f_{\texttt{CNN}}$ (the default architecture provided by \cite{matconvnet}) because it yielded better accuracy than $f_{\texttt{RFF}}$, and compare CC to the same CNN architecture with a classification layer, with all weights learned in one pass over $\cl X$ for fairness. Again CC is outperformed by the learned approach, but still achieves reasonable, non-trivial accuracy. Surprisingly, CC performs here \textit{better} on the test set than on the training set.

\begin{table}[h]
	\footnotesize
	\centering
        \scalebox{0.8}{
	\begin{tabular}{l|l|l|l|l|l|}
		\cline{2-6}
		& N                             & n                              & CNN                      & m = 250                   & m = 5000                 \\ \hline
		\multicolumn{1}{|l|}{}                          & 60000                         &                                &  $1.60 \pm 0.12$  & $17.73 \pm 1.43$ & $16.60 \pm 1.54$ \\ \cline{2-2}
		\multicolumn{1}{|l|}{\multirow{-2}{*}{MNIST}}   & \cellcolor[HTML]{C0C0C0}10000 & \multirow{-2}{*}{$28\times28\times1$}  & \cellcolor[HTML]{C0C0C0}$1.63 \pm 0.11$ & \cellcolor[HTML]{C0C0C0}$16.83 \pm 1.39$ & \cellcolor[HTML]{C0C0C0}$15.80 \pm 1.61$ \\ \hline
		\multicolumn{1}{|l|}{}  & 50000 & & $39.08 \pm 1.48$ & $71.76 \pm 1.85$ & $72.83 \pm 2.00$ \\ \cline{2-2}
		\multicolumn{1}{|l|}{\multirow{-2}{*}{CIFAR10}} & \cellcolor[HTML]{C0C0C0}10000 & \multirow{-2}{*}{$32\times32\times3$} & \cellcolor[HTML]{C0C0C0}$40.28 \pm 1.36$ & \cellcolor[HTML]{C0C0C0}$71.12 \pm 1.72$ & \cellcolor[HTML]{C0C0C0}$72.02 \pm 1.85$ \\ \hline
	\end{tabular}}
	\caption{Image datasets: train (white) and test (gray) average error rates $\pm$ standard deviation (in $\%$, 10 repetitions), for SVM and CC with $m \in \{250,5000\}$.\label{tab:exp-images}}
	\vspace{-8mm}
\end{table}

\section{Discussion and conclusion}
\label{sec:conclusion}
We proposed a very simple and flexible \textit{compressive classification} method, relying only on class sketches: accumulated random nonlinear signatures $f(\cdot)$ of the learning examples. This classifier is cheap to evaluate (\eg in low-power hardware, following ideas from~\cite{schellekens2018qckm}), involves no parameter tuning, and has an interesting interpretation: a MAP estimator inside the RKHS $\cl H_{\kappa}$ associated with the kernel $\kappa$ defined by $f$. Preliminary experimental results, relying on a basic Gaussian $\kappa$, are an encouraging proof of concept, but indicate room for improvement if the mapping $f$ (and associated kernel $\kappa$) are optimized according to the true data distribution; for example, image classification accuracy improves when $f$ is a random CNN (defining a shift-variant $\kappa$). Intuitively, $\kappa$ should be such that the Mean Maps $\kappa(\cdot,{\cl P}_k) \in \cl H_{\kappa}$ of different classes $k$ are ``well separated'' (ideally as much separated as the initial, unknown densities $\cl P_k$). This could be done by adding some a priori assumptions on the densities $\cl P_k$, or by first getting a rough estimation of them through a form of distilled sensing~\cite{haupt2011distilled}. To be reliable, compressive classification also requires precise, non-asymptotic guarantees, \eg using results from~\cite{smola2007hilbert} and~\cite{rahimi2008RFF}. 


\end{document}